\documentclass{article}
\usepackage{spconf,graphicx}
\usepackage{amsmath}
\usepackage{amssymb}
\usepackage{booktabs}
\usepackage{color}
\usepackage{url,setspace}
\usepackage[breaklinks=true]{hyperref}
\usepackage[anythingbreaks]{breakurl}
\usepackage[utf8]{inputenc} 

\newcommand{\chk}{\checkmark}
\newcommand{\fm}[1]{\footnotemark[#1]}

\usepackage[
backend=biber,
style=ieee,
maxbibnames=3,
maxcitenames=3,
doi=false,isbn=false,url=false,eprint=false
]{biblatex} 

\defbibheading{bibliography}[\refname]{}

\DeclareSourcemap{
	\maps[datatype=bibtex, overwrite=true]{
		\map{
			\step[fieldsource=booktitle,
			match=\regexp{.*Interspeech.*},
			replace={Proc. Interspeech}]
			\step[fieldsource=journal,
			match=\regexp{.*INTERSPEECH.*},
			replace={Proc. Interspeech}]
			\step[fieldsource=booktitle,
			match=\regexp{.*ICASSP.*},
			replace={ICASSP}]
			\step[fieldsource=booktitle,
			match=\regexp{.*icassp_inpress.*},
			replace={ICASSP (in press)}]
			\step[fieldsource=booktitle,
			match=\regexp{.*International.*Conference.*on.*Acoustics,.*Speech.*and.*Signal.*Processing.*},
			replace={ICASSP}]
			\step[fieldsource=booktitle,
			match=\regexp{.*International.*Conference.*on.*Learning.*Representations.*},
			replace={ICLR}]
			\step[fieldsource=booktitle,
			match=\regexp{.*International.*Conference.*on.*Machine.*Learning.*},
			replace={ICML}]
			\step[fieldsource=booktitle,
			match=\regexp{.*Automatic.*Speech.*Recognition.*and.*Understanding.*},
			replace={Proc. ASRU}]
			\step[fieldsource=booktitle,
			match=\regexp{.*Spoken.*Language.*Technology.*},
			replace={Proc. SLT}]
			\step[fieldsource=booktitle,
			match=\regexp{.*Speech.*Synthesis.*Workshop.*},
			replace={Proc. SSW}]
			\step[fieldsource=booktitle,
			match=\regexp{.*workshop.*on.*speech.*synthesis.*},
			replace={Proc. SSW}]
			\step[fieldsource=booktitle,
			match=\regexp{.*Advances.*in.*neural.*information.*processing.*},
			replace={NIPS}]
			\step[fieldsource=booktitle,
			match=\regexp{.*Workshop.*on.* Applications.* of.* Signal.*Processing.*to.*Audio.*and.*Acoustics.*},
			replace={Proc. WASPAA}]
			\step[fieldsource=publisher,
			match=\regexp{.+},
			replace={{}}]
			\step[fieldsource=month,
			match=\regexp{.+},
			replace={{}}]
			\step[fieldsource=location,
			match=\regexp{.+},
			replace={{}}]
			\step[fieldsource=address,
			match=\regexp{.+},
			replace={{}}]
			\step[fieldsource=organization,
			match=\regexp{.+},
			replace={{}}]
		}
	}
}

\title{ESPnet-TTS: Unified, Reproducible, and Integratable Open Source End-to-End Text-to-Speech Toolkit}
\makeatletter
\def\@name{
  \emph{Tomoki Hayashi}$^{1,2}$, 
  \emph{Ryuichi Yamamoto}$^3$,
  \emph{Katsuki Inoue}$^4$,
  \emph{Takenori Yoshimura}$^{1,2}$,\\
  \emph{Shinji Watanabe}$^5$,
  \emph{Tomoki Toda}$^1$,
  \emph{Kazuya Takeda}$^1$,
  \emph{Yu Zhang}$^6$, and \emph{Xu Tan}$^7$
}
\makeatother
\address{
  $^1$Nagoya University,
  $^2$Human Dataware Lab. Co., Ltd.,
  $^3$LINE Corp.,\\
  $^4$Okayama University,
  $^5$Johns Hopkins University,
  $^6$Google AI,
  $^7$Microsoft Research
}
\begin{document} 
\ninept
\maketitle
\begin{abstract}
This paper introduces a new end-to-end text-to-speech (E2E-TTS) toolkit named ESPnet-TTS, which is an extension of the open-source speech processing toolkit ESPnet.
The toolkit supports state-of-the-art E2E-TTS models, including Tacotron~2, Transformer TTS, and FastSpeech, and also provides recipes inspired by the Kaldi automatic speech recognition (ASR) toolkit.
The recipes are based on the design unified with the ESPnet ASR recipe, providing high reproducibility.
The toolkit also provides pre-trained models and samples of all of the recipes so that users can use it as a baseline.
Furthermore, the unified design enables the integration of ASR functions with TTS, e.g., ASR-based objective evaluation and semi-supervised learning with both ASR and TTS models.
This paper describes the design of the toolkit and experimental evaluation in comparison with other toolkits.
The experimental results show that our models can achieve state-of-the-art performance comparable to the other latest toolkits, resulting in a mean opinion score (MOS) of 4.25 on the LJSpeech dataset.
The toolkit is publicly available at \url{https://github.com/espnet/espnet}.
\end{abstract}


\setlength{\leftmargini}{5mm}
\begin{table*}[t]
  \begin{center}
  \vspace{0mm}
  \caption{\it Comparison with other open-source E2E-TTS toolkits, where ``$\star$'' represents under construction. Note that numbers in the table are as of October 21st, 2019.}
  \vspace{0.5mm}
  \scalebox{0.8}{\renewcommand\arraystretch{0.8}{%
  \begin{tabular}{l|ccccccc}
    \toprule
                                      & r9y9       & Kyubyong   & Rayhane    & NVIDIA   & Mozilla     & OpenSeq2Seq & ESPnet-TTS \\
    \midrule      
    Deep Voice 3                      & \chk       &            &            &          &             &             &             \\
    Tacotron                          &            & \chk       & \chk       &          & \chk        &             &             \\
    Tacotron 2                        &            &            & \chk       & \chk     & \chk        & \chk        & \chk        \\
    Transformer TTS                   &            &            &            &          &             &             & \chk        \\
    Centaur\fm{1}                     &            &            &            &          &             & \chk        &             \\
    FastSpeech                        &            &            &            &          & $\star$     &             & \chk        \\
    \midrule      
    Support multi-speaker?            & \chk       &            &            &          & $\star$     &             & \chk        \\
    Support adaptation?               &            &            &            &          &             &             & \chk        \\
    Support neural vocoder?           &         &         & \chk    & \chk & \chk     & \chk     & \chk   \\
    Support other tasks?              &            &            &            &          &             & \chk        & \chk        \\
    Provide pre-trained model?         & \chk       & \chk       &            & \chk     & \chk        & \chk        & \chk       \\
    Provide pre-trained vocoder?       &            &            &            & \chk     & \chk        &             & \chk       \\
    \midrule
    \# of stars in GitHub             & 1.1k       & 1.6k       & 1.2k       & 1.2k     & 1.4k        & 1.0k        & 1.5k        \\
    \# of supported datasets\fm{2}    & 5          & 3          & 2          & 1        & 6           & 2           & 11          \\
    \# of supported languages\fm{2}   & 3          & 1          & 8          & 1        & 8           & 8           & 11          \\
    \midrule
    Input type                        & Char / Phn & Char       & Char       & Char     & Char / Phn  & Char        & Char / Phn  \\
    Backend                           & PyTorch    & TensorFlow & TensorFlow & PyTorch  & PyTorch     & TensorFlow  & PyTorch     \\
    License                           & MIT        & Apache 2.0 & MIT        & BSD-3    & MPL 2.0     & Apache 2.0  & Apache 2.0  \\
    \bottomrule
  \end{tabular}
  }}
    \label{tab:comp2}
  \vspace{-9mm}
  \end{center}
\end{table*}

\section{Introduction}
\label{sec:intro}
Text-to-speech (TTS) is the technology to generate speech from the given input text.
TTS is an essential component for many applications such as navigation announcements on smartphones and in cars, interactive interfaces of smart assistant systems, etc.
Many researchers in speech processing communities have focused on this topic and open-source toolkits such as HTS~\cite{wg2016hts} and Merlin~\cite{wu2016merlin} have helped to accelerate their research of conventional statistical parametric speech synthesis (SPSS) systems
based on a hidden Markov model (HMM)~\cite{tokuda2013speech} and a deep neural network (DNN)~\cite{ze2013statistical}.

Recently with the success of deep learning techniques, end-to-end TTS (E2E-TTS) systems have grown in popularity and have even started replacing conventional TTS systems in production~\cite{wang2017tacotron,shen2017tacotron2,ping2017deep,li2018transformer,ren2019fastspeech}.
E2E-TTS systems do not require a complex text processing front-end depending on the language expert knowledge but also hand-annotated (e.g., phoneme alignments) corpus, and can be simply trained from pairs of text and speech.
Furthermore, it has been reported that the E2E-TTS achieves high perceptual quality comparable to professionally recorded speech~\cite{shen2017tacotron2} by incorporating neural vocoders such as WaveNet~\cite{oord2016wavenet,tamamori2017wavenet} and WaveRNN~\cite{kalchbrenner2018efficient}.
E2E-TTS is one of the most important topics in this field and various extensions including controllable and emotional E2E-TTS~\cite{hsu2019hierarchical,wang2018style} have been developed to advance the technology.

In this paper, we introduce a new E2E-TTS toolkit named ESPnet-TTS, which is an extension of the open-source speech processing toolkit ESPnet~\cite{watanabe2018espnet,karita2019tf_vs_rnn}.
The toolkit is developed for the research purpose to make E2E-TTS systems more user-friendly and to accelerate research in this field.
The toolkit not only supports state-of-the-art E2E-TTS models such as Tacotron~2~\cite{shen2017tacotron2}, Transformer TTS~\cite{li2018transformer}, and FastSpeech~\cite{ren2019fastspeech} but also provides Kaldi automatic speech recognition (ASR) toolkit~\cite{Povey_ASRU2011_2011} style recipes.
The recipe is based on the design unified with the ASR recipe and includes all of the procedures required to reproduce the results.
The toolkit provides a number of recipes for more than ten languages, which include single-speaker TTS as well as multi-speaker one and speaker adaptation.
Pre-trained models and generated samples of all of the recipes are also provided so that users can easily use it as a baseline or perform TTS demonstrations.
Furthermore, thanks to the unified design among TTS and ASR, we can easily integrate ASR functions with TTS, for example, 
ASR-based objective evaluation by character error rate (CER), which is useful to automatically find alignment failures (e.g., repetitions and deletions) in E2E-TTS~\cite{ping2017deep}.
This paper describes the basic design of the toolkit and comparative experimental evaluation with other TTS toolkits.
Subjective evaluation results verify that our best model outperforms other TTS toolkits, achieving a mean opinion score (MOS) of 4.25 on the LJSpeech dataset.

\section{Related work}

This section briefly compares ESPnet-TTS to other open-source TTS toolkits.
First, we focus on conventional TTS toolkits for SPSS: HTS~\cite{wg2016hts} and Merlin~\cite{wu2016merlin}.
HTS is one of the most popular toolkits to build an HMM/DNN-based SPSS system.
In the HMM-era, HTS has made huge contributions to developing speech synthesis technologies.
Although it can also build a simple feed-forward neural network (FNN)-based speech synthesis system, recent neural network techniques are not currently supported.
Merlin is designed for DNN-based SPSS systems and it supports various types of neural networks including mixture density networks (MDN)~\cite{zen2014deep} and recurrent neural networks (RNNs)~\cite{chung2014empirical,wu2016investigating}.
These toolkits have contributed various research topics and applications.
However, conventional SPSS systems are based on the complicated pipeline structure and each part is typically optimized separately, which may result in suboptimal performance as the whole TTS system.
On the other hand, our ESPnet-TTS is based on the end-to-end approach. 
It significantly simplifies the system structure and provides better results by optimizing the whole system in an end-to-end manner.
Furthermore, as the text processing part is integrated in the E2E model, the E2E approach allows us to construct TTS systems for various language data without expert knowledge of the target language.

Next, we focus on the comparison with other E2E-TTS toolkits.
Here, we pick up the following six toolkits available on GitHub\footnote{We selected the well-maintained toolkits which got over 1000 stars.}:
\vspace{-1pt}
\begin{itemize}
    \setlength{\itemsep}{-1pt}
    \item \textbf{r9y9}: \url{r9y9/deepvoice3_pytorch}~\cite{r9y9_deepvoice3}, 
    \item \textbf{Kyubyong}: \url{Kyubyong/tacotron}~\cite{kyubyong_tacotron}, 
    \item \textbf{Rayhane}: \url{Rayhane-mamah/Tacotron-2}~\cite{rayhane_tacotron2}, 
    \item \textbf{NVIDIA}: \url{NVIDIA/tacotron2}~\cite{nvidia_tacotron2}, 
    \item \textbf{Mozilla}: \url{mozilla/TTS}~\cite{mozilla_tts}, 
    \item \textbf{OpenSeq2Seq}: \url{NVIDIA/OpenSeq2Seq}~\cite{kuchaiev2018mixed}.
\end{itemize}
\vspace{-1pt}
\noindent Table~\ref{tab:comp2} summarizes the differences between our toolkit and the other E2E-TTS toolkits.
Compared to the other toolkits, our ESPnet-TTS provides three state-of-the-art E2E-TTS models including Tacotron~2, Transformer TTS, and FastSpeech. 
Moreover, a number of reproducible recipes are provided for more than ten languages with support for multi-speaker TTS as well as speaker adaptation techniques.
While the other toolkits provide a limited number of pre-trained models and samples, we provide them for all of the recipes, enabling the researchers to use it as a baseline system for their research and general users to play with TTS demonstrations.
Furthermore, since our ESPnet-TTS is an extension of ESPnet, both ASR and TTS recipes are based on a unified design, which allows us to easily integrate ASR functions with TTS.
For example, ASR-based objective evaluation for TTS systems and advanced research topics such as the semi-supervised learning~\cite{hayashi2018back,karita2019semi,baskar2019self,ren2019almost} can be realized by combining ASR and TTS modules in the unified framework.

\footnotetext[1]{Centaur is OpenSeq2Seq's hand-designed model.}
\footnotetext[2]{We only counted items in the official repository, not including fork ones.}

\section{Features of ESPnet-TTS}
The ESPnet-TTS consists of two main components: a library of E2E-TTS neural network models and recipes including all of the procedures to complete experiments.
The library part is written in Python using PyTorch~\cite{paszke2017automatic} as a main neural network library.
The recipes are all-in-one style scripts written in Bash and follow the Kaldi~\cite{Povey_ASRU2011_2011} style.
The following sections describe the details.

\subsection{Models}
We support three E2E-TTS models\footnote{We refer the E2E-TTS as the text to acoustic feature conversion system. The vocoder part is not included unless it is explicitly mentioned.}: Tacotron 2~\cite{shen2017tacotron2}, Transformer TTS~\cite{li2018transformer}, and FastSpeech~\cite{ren2019fastspeech}. 
The input for each model is the sequence of characters or phonemes and the output is the sequence of acoustic features (e.g. log Mel-filter bank features).

Tacotron~2 is an RNN-based sequence-to-sequence model.
It consists of a bi-directional LSTM-based encoder and a uni-directional LSTM-based decoder with location sensitive attention~\cite{chorowski2015attention}.
Different from the original Tacotron 2, we also support the forward attention w/ or w/o a transition agent~\cite{zhang2018forward}, which helps to learn diagonal attention.
As for the Transformer TTS, it adopts multi-head self-attention mechanism. 
By replacing the RNNs to the parallelizable self-attention structure, it enables faster and more efficient training while maintaining the high perceptual quality comparable to the Tacotron 2~\cite{li2018transformer}. 
For FastSpeech, it designs a feed-forward Transformer architecture for non-autoregressive generation under the teacher-student training pipeline~\cite{ren2019fastspeech}.

Furthermore, to provide multi-speaker TTS functionality, we support the use of speaker embedding as the auxiliary input for our E2E-TTS models.
We use the pre-trained x-vector~\cite{snyder2018xvector} provided by Kaldi as the speaker embedding.

\subsection{Training}
In training, we use several training criteria: the L1 loss and L2 loss for the predicted feature sequence and the weighted Sigmoid cross-entropy for the stop token sequence.
Additionally, we support the guided attention loss~\cite{tachibana2018efficiently}, which forces the attention weights to be diagonal and accelerates the learning of the diagonal attention.

Thanks to PyTorch~\cite{paszke2017automatic}, we support multi-GPU training which greatly reduces the training time, especially in the case of Transformer TTS.
This is because it needs a large batch size (e.g., $> 64$) to train Transformer stably~\cite{li2018transformer}.
However, this means that the training of Transformer requires many GPUs, which is inconvenient for light users.
To avoid this issue, we support dynamic batch making and gradient accumulation.
In dynamic batch making, the batch size is automatically adjusted according to the length of the inputs and/or outputs.
By using this scheme, we can avoid the out of memory error of GPU caused by a very long sentence, thereby improving GPU utilization.
The gradient accumulation performs backpropagation for several batches and then updates model parameters once.
This allows us to use a pseudo large batch size and as a result, we can sucessfully train Transformer with only a single GPU.

\subsection{Synthesis}
In synthesis, first we generate the log Mel-filter bank feature sequence using the trained E2E-TTS models and then use the Griffin–Lim algorithm (GL)~\cite{perraudin2013fast}, WaveNet vocoder (WNV)~\cite{oord2016wavenet,tamamori2017wavenet}, or Parallel WaveGAN (PWG)~\cite{Yamamoto2020} to generate speech from the sequence of features.
In the case of GL, we convert the sequence of log Mel-filter bank features to a linear spectrogram and then apply GL to the spectrogram.
The conversion is performed by applying the inverse Mel basis or the convolutional bank highway network GRU (CBHG) network~\cite{wang2017tacotron}.

In the case of the WNV and PWG, we use the generated log Mel-filter bank sequence as the auxiliary input of the network to generate a waveform.
We support two types of WNVs: one is that using a 16-bit mixture of logistics (MoL)~\cite{r9y9_wavenet} and the other is that using an 8-bit Softmax with the time-invariant noise shaping~\cite{hayashi_wavenet}, which can reduce perceptual noise in the high frequency band~\cite{tachibana2018invetigation}.
WNV can greatly improve the naturalness of generated speech but it requires a long time to generate. 
On the other hand, since PWG is the non-autoregressive model, it can generate much faster than the real-time while keeping the quality comparable to WNV~\cite{hayashi_pwg}.

\begin{table*}[t]
  \begin{center}
  \vspace{-3mm}
  \caption{\it Supported datasets in ESPnet-TTS, where ``$\rightarrow$'' represents down sampling, ``single'' represents the single speaker model, ``multi'' represents the multi-speaker model using a pre-trained speaker embedding, and ``adaptation'' represents the speaker adaptation which performs fine-tuning of the pre-trained model using a small amount of speech.}
  \vspace{0.5mm}
  \scalebox{0.8}{
  {\renewcommand\arraystretch{0.8}
  \begin{tabular}{l|cccccc}
    \toprule
    Dataset                            & Available languages            & Sampling rate [kHz]      & \# of speakers  & Length [hours] & Recipe type & Input type  \\
    \midrule
    Blizzard17~\cite{king2017blizzard} & En                             & 44.1 $\rightarrow$ 22.05 & 1        & 6              & Single      & Char        \\
    ARCTIC~\cite{kominek2004cmu}       & En                             & 16                       & 7        & 7              & Adaptation  & Char        \\
    CSMSC~\cite{csmsc}                 & Zn                             & 48 $\rightarrow$ 24      & 1        & 12             & Single      & Pinyin      \\
    JNAS~\cite{itou1999jnas}           & Jp                             & 16                       & 306      & 60             & Multi       & Phn         \\
    JSUT~\cite{sonobe2017jsut}         & Jp                             & 48 $\rightarrow$ 24      & 1        & 10             & Single      & Phn         \\
    JVS~\cite{takamichi2019jvs}        & Jp                             & 24                       & 100      & 30             & Adaptation  & Phn         \\
    LibriTTS~\cite{zen2019libritts}    & En                             & 24                       & 2,456    & 585            & Multi       & Char        \\
    LJSpeech~\cite{ito2017ljspeech}    & En                             & 22.05                    & 1        & 24             & Single      & Char / Phn  \\
    M-AILABS~\cite{solak2019mailabs}   & En, De, Fr, It, Es, Pl, Uk, Ru & 16                       & 30\fm{1} & 999            & Single      & Char        \\
    TWEB~\cite{tweb}                   & En                             & 12                       & 1        & 72             & Single      & Char        \\
    VAIS1000~\cite{vais1000}           & Vi                             & 16                       & 1        & 1             & Single      & Char        \\
    \bottomrule
    \end{tabular}
    \label{tab:recipe_list}
  }
  }
  \vspace{-8mm}
  \end{center}
\end{table*}

\begin{figure}[t]
    \vspace{0mm}
    \begin{center}
        \includegraphics[width=1\columnwidth]{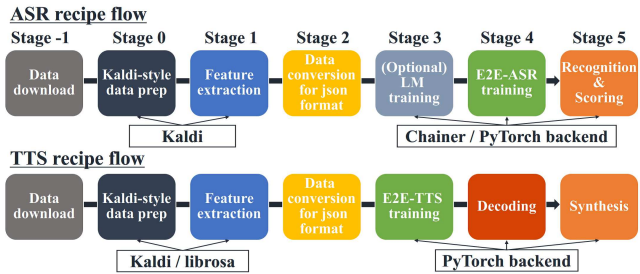}
    \end{center}
    \vspace{-6mm}
    \caption{\it Comparison between the flow of the ASR and TTS recipe.}
    \label{fig:recipe_comp}
    \vspace{-4mm}
\end{figure}

\subsection{Kaldi-style recipes}
ESPnet-TTS follows Kaldi-style data processing and provides all-in-one recipes that consist of several stages.
As you can see in Fig.~\ref{fig:recipe_comp}, the stages from $-$1 to 2 are the same in both recipes, which means we use the same data format for both the ASR and TTS models, allowing the interconversion between the ASR and TTS recipes.
The supported datasets in ESPnet-TTS are summarized in Table~\ref{tab:recipe_list}.

\subsection{Integration with ASR functions}
Thanks to the unified design, we can integrate some ASR functions with TTS.
Here we introduce some interesting functions based on the integration.

One of the common problems of E2E-TTS is that the generated speech sometimes includes the deletion and/or repetition of words in the input text due to alignment errors.
To address this issue, we provide the ASR-based objective evaluation using the CER.
Similarly to TTS, many ASR pre-trained models are provided and the TTS recipes are easily converted to ASR recipes thanks to the unification of the recipe design.
Therefore, we can evaluate the CER of generated speech using pre-trained ASR models or those trained on the converted ASR recipe and then automatically detect the deletion and/or repetition of words.

Another example based on the integration is the advanced recipes that combine ASR modules with TTS modules.
For example, we provide the recipe based on ASR-TTS cycle consistency training~\cite{baskar2019self} and semi-supervised training using ASR and TTS~\cite{karita2019semi}.
These recipes provide many tips on how to combine ASR and TTS modules and help to accelerate the further advanced research topics of the end-to-end processing.

\section{Experimental Evaluation}
\label{sec:exp}

\subsection{Experimental condition}
To demonstrate the performance of our models, we conducted experimental evaluations using the LJSpeech dataset~\cite{ito2017ljspeech}.
The dataset consists of 24 hours of English speech from a single speaker.
We used 12,600 utterances for training, 250 utterances for validation, and 250 utterances for evaluation.
To compare the performance, we trained the following six models:
\begin{itemize}
    \setlength{\itemsep}{-1pt}
    \item \textbf{Tacotron2.v2}: Tacotron~2~\cite{shen2017tacotron2} using the forward attention w/ a transition agent~\cite{zhang2018forward}
    \item \textbf{Tacotron2.v3}: Tacotron~2 trained w/ location sensitive attention~\cite{chorowski2015attention} and the guided attention loss~\cite{tachibana2018efficiently}
    \item \textbf{Transformer.v1}: Transformer TTS~\cite{li2018transformer} w/ the guided attention loss
    \item \textbf{Transformer.v3}: Transformer TTS w/ the guided attention loss and phonemes as the input
    \item \textbf{FastSpeech.v2}: FastSpeech~\cite{ren2019fastspeech} trained w/ Transformer.v1 as a teacher w/o knowledge distillation
    \item \textbf{FastSpeech.v3}: FastSpeech + Post-Net~\cite{shen2017tacotron2} trained w/ Transformer.v3 as a teacher w/o knowledge distillation
\end{itemize}
For Tacotron~2, we used the same hyperparameters in \cite{shen2017tacotron2} except for the type of the attention mechanism and the loss functions.
For Transformer TTS, we did not use the encoder Prenet in \cite{li2018transformer} since it did not provide a significant difference in our preliminary experiments.
For the phoneme conversion, we used an English grapheme-to-phoneme (G2P) library~\cite{g2pE2019}, which combines the pronunciation dictionary lookup and the sequence-to-sequence network-based predictions.
For FastSpeech, to simplify the training stage, we did not use the knowledge distillation, using the natural feature sequence as the target sequence, which is expected to affect the voice quality of FastSpeech.
We will add knowledge distillation in the next step.
The MoL-WNV was used for all of the above models, which was trained with the features extracted from natural speech.

\subsection{Objective evaluation}
First, we evaluated the E2E-TTS models using the ASR-based objective measure character error rate~(CER).
As the ASR model, we used the Transformer trained on the Librispeech dataset~\cite{panayotov2015librispeech}.
Objective evaluation results are shown in Table~\ref{tab:cer}, where Sub, Del, and Ins represent substitution, deletion, and insertion errors, respectively.
A comparison between Tacotron~2 and Transformer TTS shows that the Transformer TTS caused more deletion errors than Tacotron~2.
One of the possible reasons is that the multi-head attention is based on the dot product attention and therefore, the restriction of causality or continuity in the attention is weaker than the location-sensitive attention and the forward attention.
Therefore, it is expected that the multi-head attention based on the other types of attention mechanisms will reduce the errors.
FastSpeech.v2 achieved the best result among models, reducing all of the errors in comparison to its teacher model Transformer.v1.
This is reasonable because FastSpeech is a non-autoregressive model which, theoretically, does not cause the repetition of the word~\cite{ren2019fastspeech}.
However, FastSpeech.v3 is worse than FastSpeech.v2, especially in terms of the deletion errors.
This degradation is caused by the error of the text processing front-end which converts characters into phonemes.

\begin{table}[t]
  \begin{center}
  \vspace{-4mm}
  \caption{\it ASR-based CER results.}
  \vspace{1mm}
  \scalebox{0.8}{
  {\renewcommand\arraystretch{0.8}
  \begin{tabular}{l|cccc}
    \toprule
    Method         & Sub [\%]     & Del [\%]     & Ins [\%]     & CER [\%]     \\
    \midrule
    Groundtruth    & 0.3          & 0.7          & 0.3          & 1.3          \\
    \midrule
    Tacotron2.v2   & 0.4          & 1.0          & \ 3.6\fm{2}  & 5.0          \\
    Tacotron2.v3   & 0.5          & 1.2          & \textbf{0.3} & 2.1          \\
    Transformer.v1 & 0.6          & 1.7          & 0.5          & 2.8          \\
    Transformer.v3 & 0.5          & 1.8          & 0.5          & 2.8          \\
    FastSpeech.v2  & \textbf{0.3} & \textbf{0.9} & \textbf{0.3} & \textbf{1.6} \\
    FastSpeech.v3  & 0.4          & 1.3          & 0.4          & 2.1          \\
    \bottomrule
    \end{tabular}
    \label{tab:cer}
  }
  }
  \vspace{-8mm}
  \end{center}
\end{table}

Next, we evaluated the speed of the feature generation using the real-time factor (RTF).
In this evaluation, we used only character-based models to focus on the difference of the model architecture.
The evaluation was conducted with 16 threads of CPUs (Xeon Gold, 3.00 GHz) and a single GPU (NVIDIA TITAN V).
From the results shown in Table~\ref{tab:rtf}, all of the models can generate the features in less than RTF = 1.0 even on CPU.
Transformer TTS is slower than Tacotron~2 but FastSpeech is much faster than the other models.
Especially on GPU, FastSpeech is 30 times faster than Tacotron~2 and 200 times faster than Transformer TTS.
Since FastSpeech is a non-autoregressive model, it can fully utilize the GPU without the bottleneck of the loop processing.
Therefore, the improvement rate on GPU is higher than the other models.

\begin{table}[t]
  \begin{center}
  \vspace{-1mm}
  \caption{\it Averaged RTF results with a standard deviation.}
  \vspace{0.5mm}
  \scalebox{0.8}{\renewcommand\arraystretch{0.8}
  \begin{tabular}{l|cc}
    \toprule
    Method         & RTF on CPU        & RTF on GPU   \\
    \midrule
    Tacotron2.v2   & 0.216 $\pm$ 0.016  & 0.104 $\pm$ 0.006 \\
    Tacotron2.v3   & 0.226 $\pm$ 0.016  & 0.094 $\pm$ 0.009 \\
    Transformer.v1 & 0.851 $\pm$ 0.076  & 0.634 $\pm$ 0.025 \\
    FastSpeech.v2  & \textbf{0.015} $\pm$ \textbf{0.005}  & \textbf{0.003} $\pm$ \textbf{0.004} \\
    \bottomrule
  \end{tabular}
  }
  \vspace{-8mm}
  \label{tab:rtf}
  \end{center}
\end{table}

\subsection{Subjective evaluation}
\footnotetext[1]{The speakers in mixed data are not counted.}
\footnotetext[2]{The large insertion error is caused by the number of repetitions in a specific utterance. If we remove the sentence, it is comparable to the others.}

Finally, we conducted subjective evaluations using MOS on naturalness\footnote[3]{Audio samples used in the subjective evaluation are available at the following URL: \url{https://espnet.github.io/icassp2020-tts}}.
To verify the performance of our models, we used the following models provided by other open-source toolkits for comparison:
\begin{itemize}
    \setlength{\itemsep}{-1pt}
    \item \textbf{Merlin}~\cite{wu2016merlin}: Conventional SPSS system w/ WORLD~\cite{morise2016world},
    \item \textbf{NVIDIA}~\cite{nvidia_tacotron2}: Pre-trained Tacotron~2 w/ WaveGlow~\cite{prenger2019waveglow},
    \item \textbf{Mozilla}~\cite{mozilla_tts}: Pre-trained Tacotron~2 w/ WaveRNN~\cite{kalchbrenner2018efficient}.
\end{itemize}
Since there is no pre-trained model of the conventional SPSS system trained on the LJSpeech dataset, we manually constructed duration/acoustic models, each was based on 2 layer bidirectional LSTM RNN with 256 hidden units. The models were trained using Merlin and \url{r9y9/nnmnkwii}~\cite{ryuichi_yamamoto_2019_3270524}.
For E2E-TTS models, we selected two repositories from Table~\ref{tab:comp2} which officially provide pre-trained models.
Note that these pre-trained models were trained on the same dataset but the split of the dataset might be different (these models may have used evaluation data for training).
We used 100 sentences randomly selected from evaluation data for the subjective evaluation.
The evaluation was conducted through Amazon Mechanical Turk and the number of subjects was 101.
Each subject evaluated at least 20 samples and rated the naturalness of each sample on a 5-point scale: 5 for excellent, 4 for good, 3 for fair, 2 for poor, and 1 for bad. 
We limited subjects to people who live in the US and instructed them to use headphones and work in a quiet room.

The subjective evaluation result is shown in Table~\ref{tab:mos}\footnote[4]{The MOS of both our FastSpeech models have not yet reached the MOS of the teacher models~\cite{ren2019fastspeech}. The reason for the performance degradation from the teacher model might be because we did not use knowledge distillation. We will further investigate the reason in future work.}.
From the results, our Transformer TTS and Tacotron~2 achieved comparable performance with the other toolkit and the target speech, especially the phoneme-based Transformer (v3) outperformed all of the models.
Tacotron~2 and Transformer TTS are almost the same performance in terms of the naturalness.
It is expected that if we use phonemes as the input in Tacotron~2, the naturalness will be improved as the same as the Transformer TTS.

\begin{table}[t]
  \begin{center}
  \vspace{-4mm}
  \caption{\it MOS results with 95$\%$ confidence interval.}
  \vspace{0.5mm}
  \scalebox{0.80}{
  \renewcommand\arraystretch{0.8}
  \begin{tabular}{l|c}
    \toprule
    Method           & \hspace{5mm} MOS \hspace{5mm} \\
    \midrule
    Groundtruth      & 4.46 $\pm$ 0.05                \\
    \midrule
    Tacotron2.v2     & 4.14 $\pm$ 0.06                \\
    Tacotron2.v3     & 4.20 $\pm$ 0.06                \\
    Transformer.v1   & 4.17 $\pm$ 0.06                \\
    Transformer.v3   & \textbf{4.25} $\pm$ \textbf{0.06}                \\
    \midrule
    Merlin~\cite{wu2016merlin}     & 2.69 $\pm$ 0.09                \\
    Mozilla~\cite{mozilla_tts}     & 3.91 $\pm$ 0.07                \\
    NVIDIA~\cite{nvidia_tacotron2} & 4.21 $\pm$ 0.06                \\
    \bottomrule
    \end{tabular}
    \label{tab:mos}
  }
  \vspace{-8mm}
  \end{center}
\end{table}

\section{Summary}
\label{sec:summary}
This paper introduced a new E2E-TTS toolkit named ESPnet-TTS as an extension of open-source speech processing toolkit ESPnet.
The toolkit has been developed for the research purpose to make E2E-TTS systems more friendly and accelerate this research field.
The toolkit supports not only state-of-the-art E2E-TTS models but also various TTS recipes whose design is unified with ASR recipes, providing high reproducibility.
The experimental evaluation results demonstrated that our models can achieve state-of-the-art performance comparable to the other latest toolkits, resulting in MOS of 4.25 on the LJSpeech dataset.

In future work, we will work on the training with knowledge distillation, the support of various types of embedding such as the emotion and the accent, and further customizable network structure (e.g., multi-head attention with various attention mechanisms). 

\section{Acknowledgement}
We would like to thank Dr. Heiga Zen for his valuable comments. 

\section{References}
{
\setstretch{0.82}
\printbibliography
}
\end{document}